# Large Scale Organization and Inference of an Imagery Dataset for Public Safety

Jeffrey Liu, David Strohschein, Siddharth Samsi, Andrew Weinert
MIT Lincoln Laboratory
Email: {jeffrey.liu,david.strohschein,sid,andrew.weinert}@ll.mit.edu

*Abstract*—Video applications and analytics are routinely projected as a stressing and significant service of the Nationwide Public Safety Broadband Network. As part of a NIST PSCR funded effort, the New Jersey Office of Homeland Security and Preparedness and MIT Lincoln Laboratory have been developing a computer vision dataset of operational and representative public safety scenarios. The scale and scope of this dataset necessitates a hierarchical organization approach for efficient compute and storage. We overview architectural considerations using the Lincoln Laboratory Supercomputing Cluster as a test architecture. We then describe how we intelligently organized the dataset across LLSC and evaluated it with large scale imagery inference across terabytes of data.

## I. INTRODUCTION

With the increasing frequency and cost associated with disasters, there is a critical need to develop technology to support incident and disaster response. The Nationwide Public Safety Broadband Network (NPSBN) established and licensed by FirstNet and built and operated by AT&T is broadband network for public safety. Video applications and analytics are routinely projected as a stressing and significant service of the NPSBN. However, there is a dearth of datasets which are representative of, and tailored toward public safety operations to enable the development of computer vision capabilities optimized for public safety. This was formally identified in the NIST Public Safety Analytics R&D Roadmap [1]:

> One of the most fundamental barriers to seamless data integration is simply a lack of awareness or access to datasets that are accurate, current, and relevant to improving response.

In response, based on Weinert and Budny [2] and informed by Palen et al. [3], a video and imagery dataset of representative and operational public safety scenarios was developed by the New Jersey Office of Homeland Security and MIT Lincoln Laboratory (MITLL).

This work was performed under the following financial assistance award 70NANB17Hl69 from U.S. Department of Commerce, National Institute of Standards and Technology. DISTRIBUTION STATEMENT A. Approved for public release. Distribution is unlimited. (C) 2019 Massachusetts Institute of Technology. Delivered to the U.S. Government with Unlimited Rights, as defined in DFARS Part 252.227-7013 or 7014 (Feb 2014). Notwithstanding any copyright notice, U.S. Government rights in this work are defined by DFARS 252.227-7013 or DFARS 252.227-7014 as detailed above. Use of this work other than as specifically authorized by the U.S. Government may violate any copyrights that exist in this work.

### A. Motivation

Development of any dataset for public safety is a large combinatorial challenge, as incidents and disasters can widely vary. Additionally due to ongoing public safety operations, we envisioned the dynamism of ever growing datasets described in the First Workshop on Video Analytics in Public Safety [4]. The diversity of public safety leads to a wide ranging set of imagery and video annotations and a dataset aggregated from a variety of sources. Organizing the resulting dataset for efficient storage and compute is incredibly important as it directly influences the utilization of the dataset and promotion of computer vision capabilities to support public safety.

### B. Objectives and Contributions

The scale and scope of this dataset necessitates a hierarchical organization approach for efficient compute and storage. We overview architectural considerations using the Lincoln Laboratory Supercomputing Cluster (LLSC) as a test architecture. We then describe how we intelligently organized the dataset across the LLSC and evaluated it with large scale imagery inference across terabytes of data.

## II. TEST ARCHITECTURE AND CONSIDERATIONS

We first discuss the LLSC and recommendations by the LLSC team on how to best organize a large heterogeneous dataset for compute and storage. Similar to the YouTube-8M dataset [5], we wanted to best organize the data to enable upfront efficient machine learning.

### A. Lincoln Laboratory Supercomputing Cluster

The LLSC High-Performance Computing (HPC) systems have two forms of storage: distributed and central. Distributed storage is comprised of the local storage on each of the compute nodes and this storage is typically used for running database applications. Central storage is implemented using the open-source Lustre parallel file system[1] on a commercial storage array. Lustre provides high performance data access to all the compute nodes, while maintaining the appearance of a single filesystem to the user. The Lustre filesystem is used in most of the largest supercomputers in the world [6].

The Lustre file system consists of Metadata Servers and Object Storage Servers, which provide namespace operations and bulk IO services respectively as shown in Fig.1[2]. The various components of the system are the metadata server (MDS), object storage server (OSS), and clients.

[1]https://www.lustre.org
[2]https://www.nextplatform.com/2016/05/23/lustre-daos-machine-learning-intels-platform

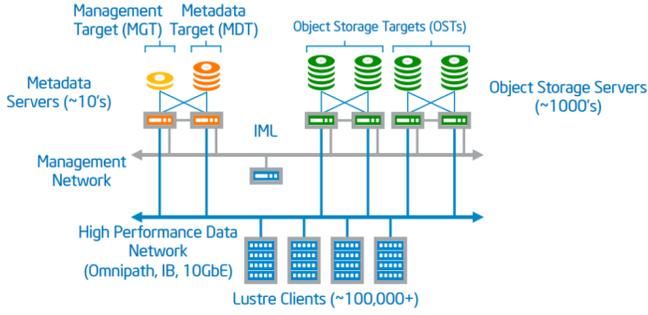

Fig. 1: Lustre architecture.

The MDS manages all name space operations for a Lustre file system. A file system's directory hierarchy and file information are contained on storage devices referred to as Metadata Targets (MDT), and the MDS provides the logical interface to this storage. OSSs provide bulk storage for the contents of files in a Lustre file system. One or more object storage servers (OSS) store file data on one or more object storage targets (OST), and a single Lustre file system can scale to hundreds of OSSs. The capacity of a Lustre file system is the sum of the capacities provided by the OSTs across all of the OSS hosts.

Applications access and use file system data by interfacing with Lustre clients. A Lustre client is represented as a file system mount point on a host and presents applications with a unified namespace for all of the files and data in the file system, using standard POSIX semantics. A Lustre file system mounted on the client operating system looks much like any other POSIX file system; each Lustre instance is presented as a separate mount point on the client's operating system, and each client can mount several different Lustre file system instances concurrently. When a client requests to open a file to the file system, it contacts the MDS with this request. The MDS checks the user authentication and the intended location of the file. Depending on the directory settings or file system settings, the MDS sends back a list of OSTs that the client can use to open the file. Once that reply is sent, the client interacts exclusively with the assigned OSTs without having to communicate with the MDS. Additionally the Lustre file system distributes segments of a file across multiple OSTs using a method called file striping. Striping has the advantage that it enables read and write operations on a file across multiple OSTs simultaneously. This can significantly increase the bandwidth when accessing a file.

### B. Data Organization for AI Applications in a HPC Environment

Small files typically use a single OST, thus serializing access to the data. Additionally, in a cluster environment, hundreds or thousands of concurrent, parallel processes accessing small files can lead to significantly large random I/O patterns for file access and results in massive amounts of networks traffic to the MDSs as described earlier. This results in increased latency for file access, higher network traffic and significantly slows down I/O and consequently causes degradation in overall application performance. This can be especially critical in AI applications that require large amounts of training data which is typically stored in small files. While this approach to data organization may provide acceptable performance on a laptop or desktop computer, it is unsuitable for use in a shared, distributed, high performace computing (HPC) system.

We store AI data in large files to take advantage of Lustre's ability to provide fast access to files. Since the block size of Lustre of is 1MB, any file created will take at least 1MB of space. In order to maximize the file I/O performance, our data is organized in large files (>100s of MB) using formats such as HDF5 or TFRecords depending on the application. If a parallel process only intends to read from these files, they are opened in read-only mode. Finally, when running distributed inference on large datasets using hundreds of parallel processes, only one process is used to get file listings or other file metadata so as to avoid excessive network traffic. This information is then broadcast to all other concurrent processes.

### C. Compute Infrastructure

The experiments described in this paper were conducted on the LLSC HPC system. This is a heterogeneous system comprising a variety of hardware platforms from AMD, Intel and NVIDIA. The cluster has compute nodes based on dual socket Haswell (Intel Xeon E5- 2683 V3 @ 2.0 GHz) processors and another single socket KNL (Intel Xeon Phi 7210 @ 1.3 GHz). Each Haswell processor has 14 cores and can run two threads per core with the Intel Hyper-Threading technology. The Haswell node has 256 GB of memory. The Intel 7210 processor has 64 cores and four hyper-threads per core and 204 GB of main memory on the compute node. Additionally, the cluster has 70 NVIDIA K80 GPUs. The GPU nodes consist of a dual socket Haswell (Intel Xeon E5-2680 v4 @ 2.40GHz) processor and two NVIDIA K80 GPUs. The K80 GPU consists of two GK210 devices with 11.44 GB of GDDR5 memory each. Thus, a process running on these compute nodes sees four GPU devices on a single compute node.

## III. DATASET

Next we overview the dataset's composition and how it is organized for archival storage, serialization, and indexing.

### A. Scale and Scope

The dataset includes images from all fifty state of the United States. It includes operational images and videos from the Civil Air Patrol (CAP), the Defense Visual Information Distribution Service (DVIDS), Massachusetts Task Force One (MA-TF1), Unmanned Robotics Systems Analysis (URSA), and the United States Geological Survey (USGS). Representative content was largely complied from Creative Commons video hosted on YouTube. A small quantity of non Creative Commons content was obtained with the permission of the content's owners. We, along with our collaborators, generated over thirty hours of video representative of some public safety scenarios. The filmed scenarios were informed by previous outreach [2]. TABLE I reports the contributions from the various data sources with Fig. 2 providing example images. The complete dataset is multiple terabytes large.

TABLE I: Imagery and video sources

| Source | Type | Approximate Scale |
|---|---|---|
| CAP | Imagery | 458,000 images |
| DVIDS | Imagery | 54 images |
| DVIDS | Video | 2 hours |
| MA-TF1 | Imagery | 9,700 images |
| MITLL + NJOHSP | Video | 35 hours |
| Massachusetts traffic cameras | Images | 150,000 |
| URSA | Video | 2 hours |
| USGS | Video | 10 hours |
| YouTube - Creative Commons | Video | 46 hours |
| YouTube - Not Creative Commons | Video | 1.5 hours |

### B. Annotations

The dataset includes human and machine generated labels. Human annotations were generated using video annotation tool from Irvine California (VATIC) [7][3] and Turkey[4] primarily on Amazon Mechanical Turk with an incentivized pricing strategy [8]. Also, machine-generated annotations from pre-trained classifiers trained on Imagenet [9], Places [10], and the Google Cloud Vision commercial classifier, provide more tags to organize and index the full dataset. Similar to the YouTube-8M dataset [5], we wanted to "remove computational barriers by pre-processing the dataset and providing state-of-art features." Details on the annotations can be found in another paper.

### C. Raw Archival

The unprocessed raw images and videos were archived as tar files split into multiple files of 4.5 GB. This allows each split file to be burned to a single layer DVD-rom for convenience while meeting the architectural considerations from Section II.

### D. HDF5 Structure and Serialization

All mission and other data is stored in hierarchical data format (HDF) file format version 5 [11] to facilitate access and processing of all the data. HDF5 is a generic format suitable for many use cases and has previously been leveraged for post-disaster imagery labels [12]. Each HDF5 file is considered an independent storage "chunk". Within each HDF5 file, data is organized in a directory structure that mirrors the original file structure from which the files were copied. This allows reading and extracting data from the HDF5 files in a manner similar to reading and writing data to a Linux file structure. Data is organized temporally; each HDF5 file contains all collected material that occurred in a particular month. Storing the data in monthly files or "chunks" facilitates downloading only the data of interest. It also enables intuitive and easy updates to the serialization as new raw data is added. For example, if a

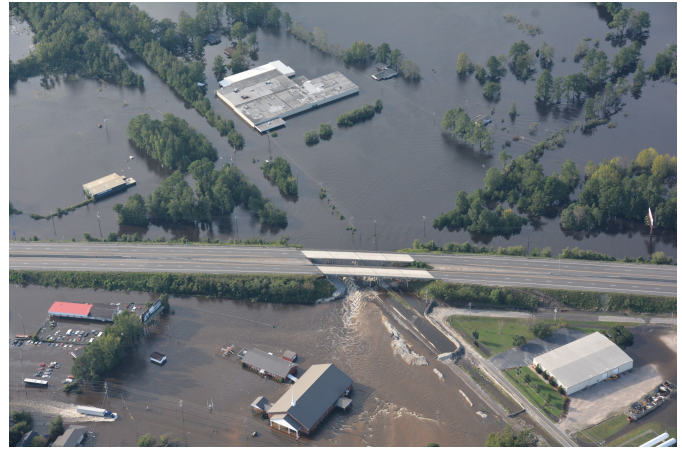
(a) Airborne perspective of flooding from the CAP

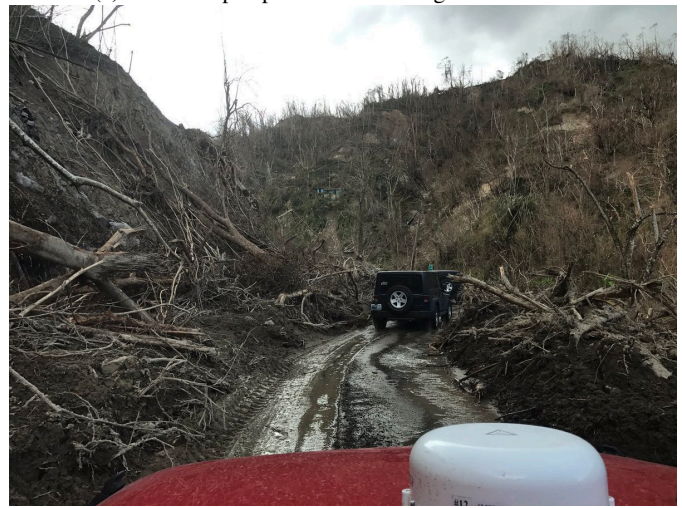
(b) Vehicle perspective of a landslide from the DVIDS

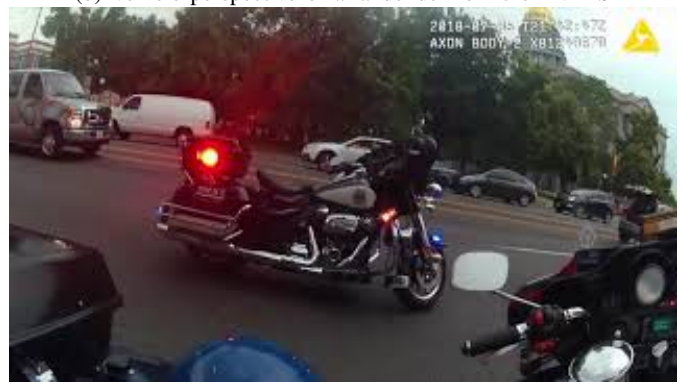
(c) Bodycam perspective from a Creative Commons video

Fig. 2: Example images included in the dataset.

---
[3]https://github.com/cvondrick/vatic
[4]https://github.com/yanfengliu/turkey

researcher is interested in a specific hurricane over a known time span, they simply need to access the files associated with known specific months. For Atlantic hurricanes, we expect 1–2 temporally organized HDF5 files will be sufficient.

This organization is outlined in Fig. 3. The organization and content of each of the HDF5 files is held within a separate entry in a NoSQL database and discussed in Section III-E. The composition of each individual HDF5 is illustrated by Fig. 4. As described previously the still imagery, video files, and associated key frames are stored in a file structure that mirrors the original organization of the data before it was copied into an HDF5 file. This file structure is just another "data group" in HDF5. In addition to this data group, there are two other user-generated files in the root group of the HDF5 file: Metadata and Annotations. These files are created to provide context and further information about the files stored in the main data group. The Metadata and Annotation files store their information as JSON data. Note, the Metadata file is a user created file and is separate from the metadata file that is created when the HDF5 file is generated as part of the normal HDF5 file creation.

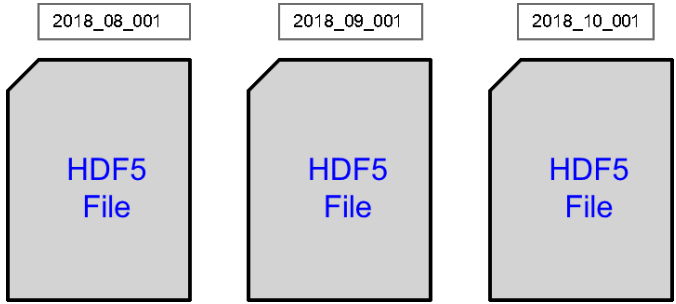

Fig. 3: Example HDF5 chunking.

### E. Data Indexing with Accumulo

The dataset is indexed using an associative array including metadata and annotations to facilitate searching through the data. Queries can be performed against the index to locate data of interest, e.g. to identify the subset of images from a specific event or location. It is implemented using the D4M paradigm with the Accumulo NoSql backend for storage [13]–[15] with Accumulo considered the one of the highest performing databases and widely used for government applications [16]. This paradigm supports sparse storage, and easy integration with analysis tools in python and matlab/octave. While the labeled data requires terabytes of storage, this indexing data can be represented by just gigabytes. Additionally there is no penalty for adding columns in Accumulo, resulting in the ability for unlimited columns and D4M doesn't require a priori knowledge of the data for ingesting or parsing, so little a priori query optimization is required. These features is critical for enabling the dataset to easily grow as new data or labels become available and meet the vision of an ever growing dataset laid out in the First Workshop on Video Analytics in Public Safety [4].

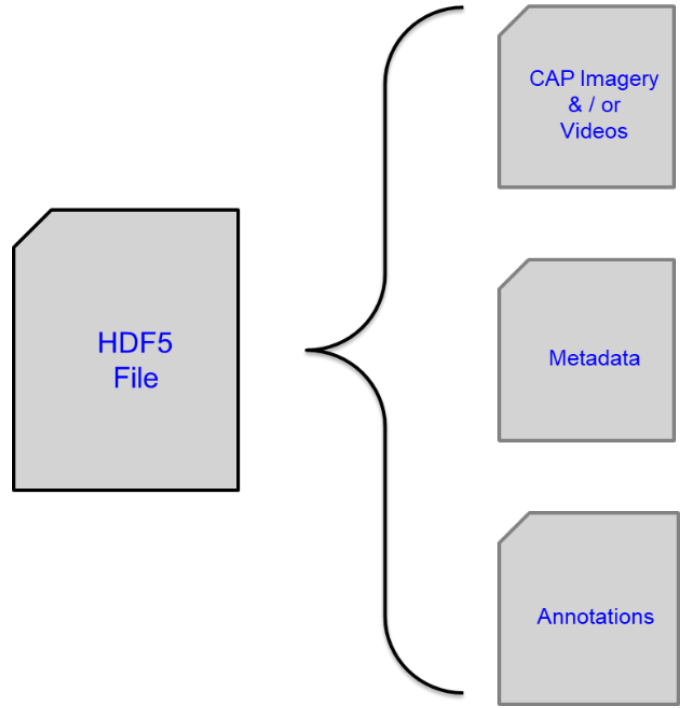

Fig. 4: Example HDF5 individual file organization.

An example structure of the index's associative array is provided in Table II. The structure of the columns and entries are given for metadata-type entries in Table III and annotation-type entries in Table IV.

- Rows are indexed by SHA1 hash (index-by-content rather than by name/location)
- Columns with hierarchical structure: *Type | Source | Field*
- Entries of associative array hold the values

|  | metadata \| {meta_source} \| {meta_field} | ... | annotation \| {anno_source} \| {anno_field} | ... |
|---|---|---|---|---|
| {file1_hash} | {meta_value} |  | {anno_value} |  |
| {file2_hash} | {meta_value2} |  | {anno_value2} |  |
| {file3_hash} | {meta_value3} |  | {anno_value3} |  |

TABLE II: Sample table structure of index associative array

## IV. INFERENCE RESULTS

This section discusses the results of large scale inference applying open source classifiers on the CAP imagery using the LLSC. In particular, we used the pretrained implementation of Inception-ResNetV2 trained on the ImageNet dataset in keras[5], and the pretrained implementation of ResNet50 trained on Places365-Standard in PyTorch[6].

| Type | Source | Field | Value (explanation) |
|---|---|---|---|
| METADATA | File | filepath | path to file location in filesystem |
| METADATA | File | HDF5 | location of file in HDF5 |
| METADATA | File | filesize | size in bytes |
| METADATA | File | dataset | high-level label categorizing the original data source, e.g. CAP, Youtube, MA-TF1 |
| METADATA | EXIF | ImageHeight | height of image in pixels |
| METADATA | EXIF | ImageWidth | width of image in pixels |
| METADATA | EXIF | GPSLatitude | latitude of image from GPS |
| METADATA | EXIF | ... | remaining EXIF specification |
| METADATA | YouTube-Info | channel | youtube channel video was uploaded to |
| METADATA | YouTube-Info | duration | length of video |
| METADATA | YouTube-Info | ... | remainder of youtube metadata fields |
| METADATA | Event | name | label for categorizing specific events, e.g. hurricane florence |
| METADATA | Event | location_state | Location information by state |
| METADATA | Event | location_kg_climate | Koppen-Geiger climate classification of location |
| METADATA | Event | ... | other event-related information, potentially sortie info, responding agencies, event characteristics |

TABLE III: Metadata-type entries

| Type | Source | Field | Value (explanation) |
|---|---|---|---|
| ANNOTATION_MACHINE | Places365 | airfield | softmax weight for label class "airfield" |
| ANNOTATION_MACHINE | Places365 | ... | softmax weight for each of the remaining classes |
| ANNOTATION_MACHINE | Imagenet | tench, Tinca tinca | softmax weight for label class "tench, Tinca tinca" |
| ANNOTATION_MACHINE | Imagenet | ... | softmax weight for each of the remaining classes |
| ANNOTATION_MACHINE | Google Cloud Vision Label Annotation | ... | returned weight of label annotations service |
| ANNOTATION_MACHINE | Google Cloud Vision Web Detection | ... | returned weight of Web Detection annotations service |
| ANNOTATION_HUMAN | Quadrant | damage | list of quadrants e.g. [NW, NE, SW, SE] where the label "damage" appears |
| ANNOTATION_HUMAN | Quadrant | ... | list of quadrants e.g. [NW, NE, SW, SE] where the label appears, for each remaining label |
| ANNOTATION_HUMAN | Point | damage | list of pixel coordinates corresponding to each instance of "damage" in the image |
| ANNOTATION_HUMAN | Point | ... | list of pixel coordinates for each instance for each remaining label class |
| ANNOTATION_HUMAN | Polygon | damage | list of lists of pixel coordinates. Each inner list corresponds to the vertices of a polygon for an instance of "damage". |
| ANNOTATION_HUMAN | Polygon | ... | list of lists of pixel coordinates. Each inner list corresponds to the vertices of a polygon of an instance of each remaining label class. |

TABLE IV: Annotation-type entries

We ran the inference task on 32 GPU nodes, using two GPUs per node. Each node processed on average 14351 images. The average runtime for the Imagenet Inception-ResNetV2 classifier was 113.25 minutes per node, and the average runtime for the Places365 ResNet50 classifier was 173.74 minutes per node. The total inference runtime for Imagenet was 60.4 node-hours, and the total runtime for Places365 was 92.7 node-hours. The measured runtime includes loading the file, and any preprocessing (rescaling and cropping) necessary to convert the image into the appropriate dimensions for the Convolutional Neural Network (CNN), in addition to the classification from the CNN. We also ran the inference task on 32 KNL CPU nodes.

## V. DISCUSSION AND FUTURE WORK

We developed and deployed a dataset organized for efficient storage and compute to enable the development of computer vision capabilities for public safety. The raw data requires terabytes of storage but the metadata and annotation indexing requires just gigabytes. In 2019, the dataset will be technology transitioned to the National Institute of Standards and Technology.

Much of the software used to develop the dataset are hosted on GitHub under BSD-2 licenses, managed by the MITLL organization, https://github.com/mit-ll, with related repositories titled "PSIAP-*."

## ACKNOWLEDGEMENT

We thank the National Institute of Standards and Technology, particularly John Garofolo and Scott McNichol. The New

---
[5]https://keras.io/applications/#inceptionresnetv2
[6]https://github.com/CSAILVision/places365

Jersey Office of Homeland Security and Preparedness have been excellent collaborators, including William Drew, Steve Talpas, Sean Gorman, Daniel Morocco, and Charles Agro. David Kovar was instrumental in obtaining operational data and informing the annotation strategy. Lastly, we thank follow MIT Lincoln Laboratory staff members, David Bestor, Chad Council, Jeremy Kepner, Dieter Schuldt, Daniel Ribeirinha-Braga, and Travis Riley.

## REFERENCES

[1] T. McElvaney, R. Felts, and M. Leh, "Public Safety Analytics R&D Roadmap," NIST, NIST Technical Note 1917, Apr. 2016. [Online]. Available: https://doi.org/10.6028/NIST.TN.1917

[2] A. Weinert and C. Budny, "Outreach to Define a Public Safety Communications Model For Broadband Cellular Video," in *2018 IEEE International Symposium on Technologies for Homeland Security (HST)*. Woburn, MA: IEEE, Oct. 2018, pp. 1–4.

[3] L. Palen, K. M. Anderson, G. Mark, J. Martin, D. Sicker, M. Palmer, and D. Grunwald, "A Vision for Technology-mediated Support for Public Participation & Assistance in Mass Emergencies & Disasters," in *Proceedings of the 2010 ACM-BCS Visions of Computer Science Conference*, ser. ACM-BCS '10. Swinton, UK, UK: British Computer Society, 2010, pp. 8:1–8:12. [Online]. Available: http://dl.acm.org/citation.cfm?id=1811182.1811194

[4] J. S. Garofolo, S. L. Garfinkel, and R. B. Schwartz, "First workshop on video analytics in public safety," NIST, NIST Pubs NISTR-8164, Jan. 2017. [Online]. Available: https://www.nist.gov/publications/first-workshop-video-analytics-public-safety

[5] S. Abu-El-Haija, N. Kothari, J. Lee, P. Natsev, G. Toderici, B. Varadarajan, and S. Vijayanarasimhan, "YouTube-8m: A Large-Scale Video Classification Benchmark," *arXiv:1609.08675 [cs]*, Sep. 2016, arXiv: 1609.08675. [Online]. Available: http://arxiv.org/abs/1609.08675

[6] J. Kepner, W. Arcand, D. Bestor, B. Bergeron, C. Byun, L. Edwards, V. Gadepally, M. Hubbell, P. Michaleas, J. Mullen, A. Prout, A. Rosa, C. Yee, and A. Reuther, "Lustre, hadoop, accumulo," in *2015 IEEE High Performance Extreme Computing Conference (HPEC)*, Sep. 2015, pp. 1–5.

[7] C. Vondrick, D. Ramanan, and D. Patterson, "Efficiently Scaling Up Video Annotation with Crowdsourced Marketplaces," in *Computer Vision âĂŞ ECCV 2010*, ser. Lecture Notes in Computer Science, K. Daniilidis, P. Maragos, and N. Paragios, Eds. Springer Berlin Heidelberg, 2010, pp. 610–623.

[8] C.-J. Ho, A. Slivkins, S. Suri, and J. W. Vaughan, "Incentivizing High Quality Crowdwork," in *Proceedings of the 24th International Conference on World Wide Web*, ser. WWW '15. Republic and Canton of Geneva, Switzerland: International World Wide Web Conferences Steering Committee, 2015, pp. 419–429, event-place: Florence, Italy. [Online]. Available: https://doi.org/10.1145/2736277.2741102

[9] O. Russakovsky, J. Deng, H. Su, J. Krause, S. Satheesh, S. Ma, Z. Huang, A. Karpathy, A. Khosla, M. Bernstein, A. C. Berg, and L. Fei-Fei, "ImageNet Large Scale Visual Recognition Challenge," *International Journal of Computer Vision (IJCV)*, vol. 115, no. 3, pp. 211–252, 2015.

[10] B. Zhou, A. Lapedriza, A. Khosla, A. Oliva, and A. Torralba, "Places: A 10 Million Image Database for Scene Recognition," *IEEE Transactions on Pattern Analysis and Machine Intelligence*, vol. 40, no. 6, pp. 1452–1464, Jun. 2018. [Online]. Available: https://ieeexplore.ieee.org/document/7968387/

[11] M. Folk, G. Heber, Q. Koziol, E. Pourmal, and D. Robinson, "An overview of the HDF5 technology suite and its applications," in *Proceedings of the EDBT/ICDT 2011 Workshop on Array Databases*, Uppsala, Sweden, 2011, pp. 36–47.

[12] K. R. Nia and G. Mori, "Building Damage Assessment Using Deep Learning and Ground-Level Image Data," in *2017 14th Conference on Computer and Robot Vision (CRV)*, May 2017, pp. 95–102.

[13] J. Kepner, W. Arcand, W. Bergeron, N. Bliss, R. Bond, C. Byun, G. Condon, K. Gregson, M. Hubbell, J. Kurz, A. McCabe, P. Michaleas, A. Prout, A. Reuther, A. Rosa, and C. Yee, "Dynamic distributed dimensional data model (D4m) database and computation system," in *2012 IEEE International Conference on Acoustics, Speech and Signal Processing (ICASSP)*, Mar. 2012, pp. 5349–5352.

[14] J. Kepner, C. Anderson, W. Arcand, D. Bestor, B. Bergeron, C. Byun, M. Hubbell, P. Michaleas, J. Mullen, D. O'Gwynn, A. Prout, A. Reuther, A. Rosa, and C. Yee, "D4m 2.0 schema: A general purpose high performance schema for the Accumulo database," in *2013 IEEE High Performance Extreme Computing Conference (HPEC)*, Sep. 2013, pp. 1–6.

[15] T. Moyer and V. Gadepally, "High-throughput ingest of data provenance records into Accumulo," in *2016 IEEE High Performance Extreme Computing Conference (HPEC)*, Sep. 2016, pp. 1–6.

[16] C. Byun, W. Arcand, D. Bestor, B. Bergeron, M. Hubbell, J. Kepner, A. McCabe, P. Michaleas, J. Mullen, D. O'Gwynn, A. Prout, A. Reuther, A. Rosa, and C. Yee, "Driving big data with big compute," in *2012 IEEE Conference on High Performance Extreme Computing*, Sep. 2012, pp. 1–6.